%% file: samplepaper.tex
\documentclass[10pt,onecolumn,letterpaper]{article}

\usepackage{graphicx}
\usepackage{times}
\usepackage{epsfig}
\usepackage{graphicx}
\usepackage{amsmath}
\usepackage{amssymb}
\usepackage{algpseudocode}
\usepackage{algorithm}
\usepackage{hyperref}
\usepackage{url}
\usepackage{cite}
\usepackage{xcolor}
\usepackage{tabularx}
\newcolumntype{E}{ >{\centering\arraybackslash} m{2cm} }
\newcolumntype{X}{ >{\centering\arraybackslash} m{3cm} }
\newcolumntype{V}{ >{\centering\arraybackslash} m{3.5cm} }

\algnewcommand\algorithmicforeach{\textbf{for each}}
\algdef{S}[FOR]{ForE1ach}[1]{\algorithmicforeach\ #1\ \algorithmicdo}

\begin{document}
\title{CORAL8: Concurrent Object Regression for Area Localization in Medical Image Panels}

%
%
\author{Sam Maksoud\footnote{\tt\small s.maksoud@uqconnect.uq.edu.au; arnold.wiliem@ieee.org; \{k.zhao1, t.zhang, lin.wu\} @uq.edu.au; lovell@itee.uq.edu.au}, Arnold Wiliem, Kun Zhao, Teng Zhang, \\
 Lin Wu and Brian C. Lovell \\
~\\
The University of Queensland, School of ITEE, QLD 4072, Australia\\
}

\maketitle              % typeset the header of the contribution
\input{abstract}

\input{introduction}

\input{dataset}

\input{proposed_method}

\input{experiments}

\input{discussion}

\section{Acknowledgements}
This project has been funded by \href{https://www.snp.com.au/about-us/about-sullivan-nicolaides-pathology/}{Sullivan Nicolaides Pathology} and the \href{https://www.arc.gov.au/}{Australian Research Council (ARC) Linkage Project} [Grant number LP160101797].

%\input{conclusions}


{\small
\bibliographystyle{splncs03}
\bibliography{bib}
}

\end{document}

\title{CORAL8: Concurrent Object Regression for Area Localization in Medical Image Panels}
\titlerunning{CORAL8}
\maketitle              % typeset the header of the contribution

\input{appendix}

\bibliographystyle{splncs03}
\bibliography{sup_bib}

%% file: abstract.tex
\begin{abstract}

This work tackles the problem of generating a medical report for multi-image panels. We apply our solution to the Renal Direct Immunofluorescence (RDIF) assay which requires a pathologist to generate a report based on observations across the eight different WSI in concert with existing clinical features. To this end, we propose a novel attention-based multi-modal generative recurrent neural network (RNN) architecture capable of dynamically sampling image data concurrently across the RDIF panel. The proposed methodology incorporates text from the clinical notes of the requesting physician to regulate the output of the network to align with the overall clinical context. In addition, we found the importance of regularizing the attention weights for word generation processes. This is because the system can ignore the attention mechanism by assigning equal weights for all members. Thus, we propose two regularizations which force the system to utilize the attention mechanism. Experiments on our novel collection of RDIF WSIs provided by a large clinical laboratory demonstrate that our framework offers significant improvements over existing methods.

\end{abstract}

%% file: introduction.tex
\section{Introduction}
\label{section:introduction}
Automatic image captioning~\cite{xu} is an important topic in the medical research area as it frees pathologists from manual medical image interpretation and reduces the cost significantly~\cite{EXPENSE}. Typically this involves conditioning a recurrent neural network (RNN) on image features encoded by a convolutional neural network (CNN). This method has shown great promise in non-specific image captioning tasks but has not generalized well to the complex domain of medical images~\cite{mdnet, compare}.\\
\\
A common solution is to employ a pathologist to annotate training data\cite{mdnet}. However, even with access to annotated image features and medical reports, the overall clinical context is still important in medical image interpretation. This is because certain pathologies may be morphologically indistinguishable. One such case is the differential diagnoses of immunotactoid glomerulonephritis and diabetic nephropathy. In the Renal Direct Immunofluorescence (RDIF) assay, both conditions can present with linear accentuation of the glomerualar basement membrane for IgG; the significance of this pattern cannot be determined without clinical confirmation/exclusion of diabetes mellitus~\cite{CLIN_CON}. For this reason, image captioning models conditioned solely on image data may not be the ideal solution for tasks in the medical domain. \\
\\
\noindent
The second major challenge of image captioning task in medical domain is correlating information from multiple images. For example, a pathologist must interpret a set of 8 different renal biopsy sections of the same patient to report the RDIF assay. Several methods have been proposed to enable captioning of multiple images by assuming images in the set exhibit temporal dependence \cite{VID1} or contain multiple views/instances of the same object\cite{compare}. These assumptions are unsuitable for the multi-object temporally independent RDIF set. We refer to this problem as the ordered set to sequence problem: it must be an ordered set to preserve the identity of the antibody in the RDIF panel.\\
\\
To address the problems outlined above, we describe a novel framework to overcome the clinical context bias and generate a RDIF medical report. The contributions of this paper are listed as follows:

\begin{enumerate}
  \item To our knowledge, we are the first work to solve the ordered set to sequence problem by proposing a novel attention based architecture which provides concurrent access to all images at each step and models the clinical notes as priors to regulate clinical contexts.
  \item We also introduce two novel regulators, Salient Alpha (SAL) and Time Distributed Variance (TDVAR), to discourage uniform attention weights.
  \item Finally, we will release a novel RDIF dataset\footnote{\url{https://github.com/cradleai/rdif}} with quantitative baseline results provided using metrics from BLEU~\cite{BLEU}, ROUGE~\cite{ROUGE} and METEOR~\cite{METEOR}.
  
\end{enumerate}

%% file: dataset.tex
\section{Renal Direct Immunoflourescence Dataset}
\label{dataset}

The novel RDIF dataset used in this paper was assembled from routine clinical samples in collaboration with \href{https://www.snp.com.au/about-us/about-sullivan-nicolaides-pathology/}{Sullivan Nicolaides Pathology}; a subsidiary of \href{https://www.sonichealthcare.com/services/}{Sonic Healthcare Limited}. To prepare the RDIF slides, eight separate sections of renal biopsy tissue are treated with fluorescein isothiocyanate (FITC) conjugated antibodies to one of either IgG, IgA, IgM, Kappa, Lambda, C1q, C3 or Fibrinogen antibodies. The dataset comprises of 144 patient samples split into 99 training, 15 validation and 30 test sets.  Each sample contains the eight WSI's, the clinical notes of the requesting physician, and the medical report.

%% file: proposed_method.tex
\section{Proposed Method - CORAL8}
\subsection{Architecture}
\label{CORAL8:Architecture}
As illustrated in Fig.~\ref{section:introduction}, the main aim of CORAL8 is to generate a medical report, $\mathcal{R}$ from an ordered set of images $\mathcal{I}$ and clinical notes $\mathcal{Q}$. To this end, we train a sentence generator $\phi_{s}$ which receives a report context vector and local image features vector as input and generates sequence of words. There are several desirable properties that we enforce in the sentence generator:
\begin{enumerate}
\item It must generate coherent sentences;
\item The generated sentences must be in concert with the clinical contexts described in the clinical notes;
\item The attention mechanism must produce diverse representations of local image features for the report generation. The mechanism must also ensure that local features from each image in the set are equally represented in the generative sequence.
\end{enumerate}
\label{section:CORAL8}
\begin{figure}
\begin{center}
   \includegraphics[width=\linewidth]{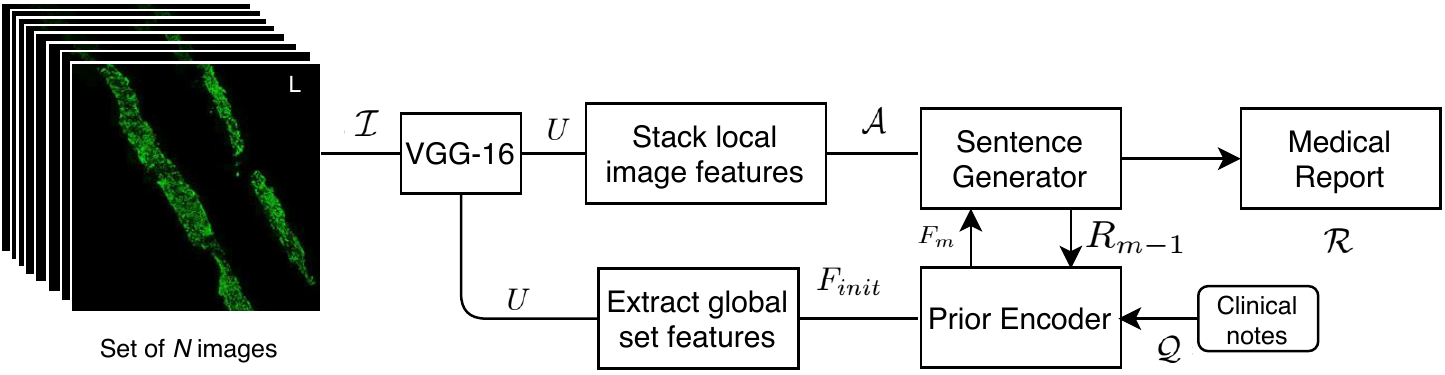}
\end{center}
   \caption{This image illustrates the framework of the proposed CORAL8 Architecture.}
\label{fig:CORAL8}
\end{figure}
\noindent
To generate coherent sentences, we train a neural network called the prior encoder $\phi_{p}$ which extracts context features $F_m$; where $m$ is the index of the sentence. The context features are a joint representation of; (1) previous sentence features; and (2) previous context features. To encourage agreement between the generated report and the clinical context, we feed the prior encoder with; (1) clinical notes features and (2) global image set features. This can be interpreted as forming a general impression of the image features with respect to the clinical context. The context features will then be fed into the sentence generator. Finally, to ensure that the model attends to each image in the set, we use a regulated attention mechanism to generate a dynamic local image features conditioning vector $L_{t}$ used to generate the next word in the sentence. We describe these components in detail below. \\
\\
\textbf{Image Encoder} encodes an ordered set of images $\mathcal{I} = \{ I_{1}, .., I_{N} \} $  into the set of local image features $\mathcal{A}$ and the global image features $F_{init}$. More specifically, we first extract the $14$x$14$x$512$ features $U$ from the 4th max-pool layer of the pre-trained VGG-16 network~\cite{vgg}. Then the extracted features $U$ are concatenated and flattened to compute the local image representations,  $\mathcal{A} = \{ A_{1},...,A_{a} \}, A_{i} \in \mathbb{R}^{N \text{x} d}$  where $a = 196$ and $d = 512$ are the dimensions of the flattened image and feature channels respectively. Meanwhile, in order to produce a global representation for $\mathcal{I}$, we apply an FC layer to $U$ to extract $F_{init}$ with $1$x$H$ fixed dimensions.  Then, $\mathcal{A}$ and $F_{init}$ are fed into the following attention and context encoder networks.\\
\\
\textbf{Prior Encoder} extracts the context vector $F_{m}$ to represent features from the clinical notes and previously generated sentence $R_{m-1}$. More specifically, for each sentence we first use a word embedding to produce fixed vector representations $\mathcal{S} = \{s_{1},...,s_{C}\},s_{t} \in \mathbb{R}^{V \text{x} E}$ and $\mathcal{Q}=\{q_{1},...,q_{C}\},q_{i} \in \mathbb{R}^{V \text{x} E}$  for the set of words in the previous sentence and clinical notes respectively; where, $V$ is the size of the vocabulary, $C$ is the number of words in each sentence, $E=512$ is the word embedding space and $q_{i}$ and $s_{t}$ are both  1-of-$V$ encoded words. $R_{m-1}$ is then fed into a bidirectional LSTM~\cite{LSTM} and followed by an FC layer to encode a fixed $1$x$H$ representation $J_{m}$. We then concatenate $J_{m}$ and $F_{m-1}$ and apply an FC layer to encode $F_{m}$ with $1$x$H$ fixed dimensions. At $m = 0$, the output of the image encoder $F_{init}$ and the embed clinical notes $\mathcal{Q}$ are used in place of $F_{m-1}$ and $R_{m-1}$ respectively.\\
\\
\textbf{Sentence Generator} is an RNN that generates a sentence $\hat{R}_{m}$ word by word conditioned on the outputs of image encoder $\mathcal{A}$ and prior encoder $F_{m}$. After a sentence is generated, the prior encoder receives it as input to generate $F_{m+1}$ which is then fed back to the sentence generator to generate the next sentence: this is repeated until the entire medical report is generated. More specifically, the attention network first computes a probability distribution over $\mathcal{A}$ using the deterministic soft attention methods described in ~\cite{xu} to compute $\kappa_{ti}$.  $\kappa_{ti}$ is then used to compute the weighted inter-image features vector $\mathcal{Z}_{t} = \phi(\{A_{i}\},\{\kappa_{i}\}) = \sum^a_{i} \kappa_{i}A_{i}$.  The network then computes a second probability distribution $\alpha_{ti}$ in the same way over $\mathcal{Z}_{t} =   \{z_{1},...,z_{N}\},z_{i} \in \mathbb{R}^{d}$. The second soft attention weighted inter-image feature vector $L_{t}= \phi(\alpha(\{z_{i}\},\{\alpha_{i}\}) = \sum^N_{i} \alpha_{i}z_{i}$ serves as the $1$x$d$ local features conditioning vector used to generate $s_{t+1}$. $L_{t}$ and $h_{t}$ are then passed through to a visual sentinel. The visual sentinel multiplies $L_{t}$ and $F_{m}$ by gating scalars $\beta_{L}$ and $\beta_{F}$. Both gating scalars; are computed as follows\\
\begin{equation}
\label{eq:beta_gate}
\beta_{x} = Sigmoid(w_{x}h_{t} + b_{x})
\end{equation}
where $w_{x}$ and $b_{x}$ are hyper-parameters to be learned by 
the network. This allows the network to judge the importance of $L_{t}$ and $F_{m}$ features when generating $s_{t+1}$. 
$\beta_{L}L_{t}$, $\beta_{F}F_{m}$ and $s_{t}$ are concatenated and fed into a LSTM. A deep multilayer perceptron output layer (MLP)~\cite{deepoutput} then computes $\mathcal{X}_{prob} = \{ p_{i}, .., p_{V} \}$; the probability distribution over the vocabulary of V words using the $1$x$H$ hidden state $h_{t}$ output of the LSTM. Specifically, MLP takes $s_{t}$, $L_{t}$, $h_{t}$ and $F_{m}$ as input and computes the probability distribution for $s_{t}$ as:\\
\begin{equation}
\label{eq:output}
p(S_{t} | S_{t-1}, L_{t}, F_{m}) = \tanh(W_{v}(S_{t-1} + W_{h}h_{t} + W_{L}L_{t} + W_{F}F_{m}))
\end{equation}
Where $W_{v} \in \mathbb{R}^{V\textbf{x}E}$, $W_{h} \in \mathbb{R}^{H\textbf{x}E}$, $W_{L} \in \mathbb{R}^{E\textbf{x}d}$, $W_{f} \in \mathbb{R}^{E\textbf{x}H}$ are all parametrized by the neural network. We then apply an $arg max$ function to $\mathcal{X}_{prob}$ to generate the next word in the sentence. This process is repeated for all words in the sentence.\\

\subsection{Attention Regularization}
\noindent
When weights of the attention mechanism are not regularized, there is a possibility that the network will assign each data point equal attention weights. To this end, we apply a set of regularizations on the attention weights to enforce the network not to choose all the data as its base in making the decision. \\
\\
\noindent
\textbf{Xu's \textit{et al.} -- } We first apply the regularization proposed by Xu \textit{et al.} ~\cite{xu} to ensure that all the attention weights sum to one for both temporal direction and spatial direction for local image features extraction. More specifically, Xu \textit{et al.} encourage a doubly stochastic property on the attention matrix which contains attention weights for local image feature extraction at each time step. The loss function for Xu \textit{et al.} is defined as:
\begin{equation}
\label{eq:xu}
C_{xu} = \sum^{N}_{i}(1-\sum^{C}_{t}\alpha_{ti})^2
\end{equation}
\noindent
\textbf{Salient Alpha (SAL) -- } The aim for SAL regularization is to increase the distance between the maximum weight and the average weight to force the network to be highly selective when attending to image regions.
We define SAL as follows.
\begin{equation}
\label{eq:SAL}
C_{SAL} = \frac{1}{C} \sum^{C}_{t=0}\left(\frac{max_i(\alpha_{ti})-mean_i(\alpha_{ti})}{mean_i(\alpha_{ti})}\right)
\end{equation}
Where $max_{i}$ is the maximum value and $mean_{i}$ is the mean along the column axis. \\
\\
\noindent
\textbf{Time Distributed Variance (TDVAR) -- }  The TDVAR aims to increase the variance of the attention weights. This will enforce the network to assign different attention weights for each generated word and enforces high variability in the attended features when generating the text sequence. We define TDVAR as follows:
\begin{equation}
\label{eq:TDVAR}
C_{TD} = \frac{1}{N} \sum^{N}_{i=0}\left(\frac{std_{t}(\alpha_{ti})}{mean_{t}(\alpha_{ti})}\right)
\end{equation}
Where $std_{t}$ is the standard deviation and $mean_{t}$ is the mean along the row axis of $\alpha_{ti}$. We then combine these three regularization terms to produce $C_{alpha}$ as follows:
\begin{equation}
\label{eq:alphareg}
C_{alpha} = \lambda_{1} C_{xu} +  \frac{\lambda_{2}}{\max(\delta, C_{SAL})} + \frac{\lambda_{3}}{\max(\delta, C_{TD})}
\end{equation}
\noindent
where $\lambda_{1}$, $\lambda_{2}$ and $\lambda_{3}$ are hyperparameters to scale the representation of each term. $\delta$ is used to avoid zero division and exploding gradients in the initial training steps.

\subsection{Training protocol}
We use random initialisation for neural network weights and zero initialisation for biases. The embedding matrix $\mathbb{R}^{V \text{x} E}$ is randomly initialised with values between $-1$ and $+1$. During training, the input to the network is $\mathcal{D}_{t} = \{\mathcal{I}, \mathcal{R},  \mathcal{Q}\}^{D_{t}}_i=0$. The cost function used to train the network is as follows;
\begin{equation}
\textbf{Cost} = -\log(P(\mathcal{R}|\mathcal{I}\cap \mathcal{Q})) + C_{alpha}
\end{equation}
We update the gradients of the network using truncated backpropgation through time (TBTT) with $\tau= 2m$~\cite{tbtt} and ADAM optimisation~\cite{adam} \textit{ı.e.} The error is computed over the generated sentence, and the prior encoded previous sentence of lengths $m$. We implement the norm clipping strategy of~\cite{clip} to stabilize the network and prevent exploding gradients. During inference, the inputs to the network are  $\mathcal{D}_{in} = \{\mathcal{I}, \mathcal{Q}\}^{D_{in}}_{i=0}$. 
\textbf{NEWLINE} tokens serve as the initial word inputs to the sentence generator; which generates the sentence word by word. Sentences are then generated one by one until the entire medical report sequence is complete.

%% file: experiments.tex
\section{Experiments and Results}
\label{section:experiments}
\begin{table}
\scriptsize
\center
\caption{This table provides the quantitative evaluations metrics for the machine generated texts.}
\label{RES}
\begin{tabular}{|c|c|c|c|c|c|c|}
\hline
Architecture & BLEU-1 & BLEU-2 & BLEU-3 & BLEU-4 & ROUGE & METEOR \\
\hline
CORAL8 & \textbf{0.49} & \textbf{0.35} & \textbf{0.28} & \textbf{0.23} & \textbf{0.39} & \textbf{0.31}\\
Recurrent Attention~\cite{compare} & 0.36 & 0.26 & 0.21 & 0.17 & 0.30 & 0.29\\
Xu \textit{et al.}~\cite{xu} & 0.44 & 0.30 & 0.23 & 0.19 & 0.35 & 0.30 \\
VANNILA & 0.31 & 0.19 & 0.12 & 0.07 & 0.30 & 0.29\\
CORAL8 w/o clinical notes & 0.44 & 0.29 & 0.22 & 0.17 & 0.34 & 0.29\\
CORAL8 w/o TDVAR & 0.42 & 0.22 & 0.17 & 0.17 & 0.37 & 0.26\\
CORAL8 w/o Xu's regularization & 0.42 & 0.30 & 0.23 & 0.19 & 0.36 & 0.29\\
CORAL8 w/o SAL & 0.41 & 0.29 & 0.22 & 0.18 & 0.36 & 0.29\\
CORAL8 w/o attention regularization& 0.44 & 0.31 & 0.25 & 0.20 & 0.37 & 0.28\\
\hline
\end{tabular}
\end{table}
\noindent
We prepare the dataset by resizing all images to  224x224x3 pixels in order to make use of a VGG-16 network pre-trained on ImageNet~\cite{imagenet}. All words from the medical reports and clinical notes are tokenized and we replace any word that occurs less than two times in the dataset with a special \textbf{UNK} token. This creates the vocabulary of 596 words. \textbf{NEWLINE} and \textbf{EOS} tokens are added to every sentence to indicate the start and end of the sentence respectively. Each sentence is padded to a fixed length of 40 word with \textbf{NULL} tokens. Finally, each medical report is padded to a fixed length of seven sentences.\\
\\
We trained our model for 30 epochs using a learning rate of 0.001, $\lambda_{1}=1$, $\lambda_{2}=0.5$, $\lambda_{3}=0.5$ and $\delta = 0.001$. Then, the performance was evaluated using BLEU~\cite{BLEU}, ROUGE~\cite{ROUGE}, and METEOR~\cite{METEOR}. We use an implementation of ~\cite{compare} to serve as the baseline for our experiment; using the set of eight RDIF images as input instead of the pair of radiology images. We refer to this method as \textbf{Recurrent Attention}. The alpha regularization method proposed in~\cite{xu} applied to our CORAL8 model is also compared as a baseline. We conduct ablation studies to evaluate the contributions of each proposed component to the overall performance of our model. To determine the significance of clinical note features, we train a model that omits the initial prior encoder step and uses $F_{init}$ to represent the context vector for the first sentence \i.e. $F_{0} = F_{init}$. Examples of generated reports with and without clinical notes can be found in Table.~\ref{GEN} To asses the effects of each alpha regularization term; we train a model that omits it from the cost function. We also include a vanilla implementation where the sentence generator consists of LSTM conditioned only on $S_{t}$ and $F_{init}$. The quantitative results for all models are provided in Table ~\ref{RES}.

\begin{table}
\scriptsize
\center
\caption{This table contains abridged examples of machine generated reports. Due to space constraints, we only include the final sentence of the report as it contains the overall impressions. Unabridged  examples are included in the supplementary materials. Key words are highlighted red}
\label{GEN}
\begin{tabular}{|E|X|X|V|} \hline
 & \textbf{Example 1} & \textbf{Example 2} & \textbf{Example 3} \\ \hline

\textbf{Clinical Notes} & Renal biopsy. For IF and histology. Creatinine 250. Proteinuria, haematuria, \textbf{\color{red}{suspected IgA nephropathy}} &  histopathology, IF.  Renal failure.  Urine protein -/blood positive. \textbf{\color{red}{ANCA positive. ?ANCA vasculitis.}}  ? Crescentic necrotising GN & Renal Bx.  Diabetic.  \textbf{\color{red}{Hypertensive}}.  eGFR 23.  Proteinuria.\\ \hline
\textbf{Ground Truth} & iga UNK oxford classification s1 t2 UNK and UNK are UNK due to UNK tissue UNK &  pauci immune anca related focal segmental necrotising glomerulonephritis & arterionephrosclerosis with UNK UNK interstitial fibrosis and hypertensive vascular disease glomerulomegaly consistent with grade 1 diabetic glomerulopathy widespread low grade tubular epithelial injury with some atn\\
\hline
\textbf{CORAL8\newline \emph{with}\newline clinical notes} &  iga nephropathy oxford classification m0 t0 &  focal segmental necrotising and crescentic pauci immune glomeruli nephritis & acute on chronic thrombotic microangiopathy tma UNK a history of UNK hypertension\\
\hline
\textbf{CORAL8 \newline \emph{without}\newline clinical notes} &  fsgs & there is equivocal reactivity for igg igm c3 and c1q of the glomeruli with a similar intensity. & there is no.\\ 
\hline
\end{tabular}
\end{table}

%% file: discussion.tex
\section{Discussion and future direction}
\label{section:discussion}

Table~\ref{RES} shows that removing any component of the alpha regulation term will result in decreased performance. This suggests that the introduced SAL, TDVAR and clinical notes all contribute to the final performance. Insights into how the clinical notes may be improving results can be inferred from Table~\ref{GEN}. \\
\\
The first example refers to a case of IgA nephropathy; oxford classifcation S1 T1. S1 indicates that some glomeruli are segmentally sclerosed; this is both a feature of IgA nephropathy and focal segmental glomerular sclerosis (FSGS). The model without clinical notes concluded the image was FSGS, but the presence of \emph{suspected IgA nephropathy} in the clinical notes resulted in the proposed model predicting IgA nephropathy in the generated report. In the second example, the proposed model accurately predicts pauci immune glomeruli nephritis. This condition is often referred to as Anti-neutrophil cytoplasmic antibody (ANCA) associated vasculitis, due to its strong association with ANCA antibodies~\cite{ANCA}. The presence of \emph{ANCA positive ?ANCA vasculitis} in the clinical notes suggests that the proposed model is capturing the associations between the clinical context and the pathologist impressions of the RDIF assay. Example 3 in Table~\ref{GEN} illustrates an incorrect impression generated by the CORAL8 model. Despite being incorrect, the example illustrates how conditions with similar clinical features are modelled by the network. The underlying pathophysiology for both the ground truth condition (arterionephrosclerosis) and the predicted condition (thrombotic microangiopathy) can be due to the clinically observed hypertension~\cite{ati,tma}. This suggests that the distributions are similar for pathologies that present with shared clinical features; indicating a degree of semantic understanding. \\
\\

%% file: appendix.tex
\section{Supplementary Materials}
\label{section:Appendix}

\begin{table}
\caption{This table shows the results of our validation test to asses the credibility of our implementation of the recurrent attention model described in~\cite{compare}. Here we provide a comparison on the same publicly available chest x-ray collection~\cite{openi}}
\label{RADIO_COMPARE}

\begin{tabular}{|c|c|c|c|c|c|c|}
\hline
Model & BLEU-1 & BLEU-2 & BLEU-3 & BLEU-4 & ROUGE & METEOR \\
\hline
OUR RECURRENT ATTENTION MODEL & \textbf{0.46} & 0.31 & 0.22 & 0.16 & \textbf{0.38} & \textbf{0.28}\\
QUOTED PERFORMANCE~\cite{compare} & \textbf{0.46} & \textbf{0.36} & \textbf{0.27} & \textbf{0.20} & 0.37 & 0.27\\
\hline
\end{tabular}
\end{table}

\begin{table}
\caption{This table shows examples of complete reports generated by the CORAL8 model with and without clinical notes as inputs. Key words in the clinical notes are highlighted in red}
\begin{tabular}{|p{1.5cm}|p{4cm}|p{4cm}|p{4cm}|} \hline

\textbf{Clinical Notes} & Renal biopsy. For IF and histology. Creatinine 250. Proteinuria, haematuria, \textbf{\color{red}{suspected IgA nephropathy}} &  histopathology, IF.  Renal failure.  Urine protein -/blood positive. \textbf{\color{red}{ANCA positive. ?ANCA vasculitis.  ? Crescentic necrotising GN}} & Renal Bx.  Diabetic.  \textbf{\color{red}{Hypertensive}}.  eGFR 23.  Proteinuria.\\ \hline
\textbf{Ground Truth} & The sample submitted for immunofluorescence study comprises renal cortex including up to six (6) glomeruli, five (5) of which show global sclerosis of obsolescent type and the remaining glomerulus exhibiting segmental sclerosis. There are no significant reactions in any tissue compartment for IgG and fibrinogen. There is global mesangial deposition of IgA ($3+$) and C3 ($3+$) with IgM ($2+$), C3 and C1q ($1+$) present in the sclerotic segment. Kappa and lambda light chains are of similar ditrsibution to IgA ($both 2+$). iga UNK oxford classification s1 t2 UNK and UNK are UNK due to UNK tissue UNK & the sample submitted for immunofluorescence comprises renal cortex including twenty two UNK glomeruli nine 9 of which show global sclerosis. there is focal segmental trace reaction for iga and igm and focal segmental c3 $1+$. there are no significant reactions in any tissue compartment for igg c1q kappa and lambda light chains and fibrinogen. pauci immune anca related focal segmental necrotising glomerulonephritis & the sample submitted for immunofluorescence comprises cortex that contains three 3 glomeruli one 1 of which is converted into a shrunken UNK. the functional glomeruli are UNK. reactions against igg iga igm c3 c1q kappa and lambda light chains and fibrinogen are all negative.arterionephrosclerosis \color{black}{with UNK UNK interstitial fibrosis and }hypertensive vascular disease glomerulomegaly\color{black}{ consistent with grade 1 diabetic glomerulopathy widespread low grade tubular epithelial injury with some atn}\\
\hline
\textbf{CORAL8\newline \emph{with}\newline clinical notes} & the sample submitted for immunofluorescence study comprises renal cortex including seven 7 glomeruli one 1 of which shows global sclerosis of obsolescent type. there are no significant reactions in any tissue compartment for igg iga igm c3 c1q kappa and lambda light chains and fibrinogen. reactions against igg iga igm c3 c1q kappa and lambda are negative. iga nephropathy oxford classification m0 t0 & sections show one core of renal cortex which includes up to 10 glomeruli in a single level of which up to 2 are globally sclerosed. there is linear accentuation of glomerular and tubular basement membranes with igg. focal segmental necrotising and crescentic pauci immune glomeruli nephritis & the sample submitted for immunofluorescence comprises cortex that contains up to six glomeruli. there is no global segmental sclerosis. there are no significant reactions in any tissue compartment of igg iga igm c3 c1q kappa and lambda light chains and fibrinogen. acute on chronic thrombotic microangiopathy tma UNK a history of UNK hypertension\\
\hline
\textbf{CORAL8 \newline \emph{without}\newline clinical notes} &  the sample submitted for immunofluorescence comprises cortex that contains more than ten 10 glomeruli. there is focal segmental mesangial and peripheral capillary wall positivity for igm 1+. reactions against igg iga UNK c1q kappa and lambda light chains and fibrinogen are all negative. UNK UNK UNK of red fsgs & sections show one core of renal cortex which includes up to 5 glomeruli in a single level none of which is globally sclerosed. reactions against igg iga c3 c1q kappa and lambda UNK chains and fibrinogen are all negative. there is equivocal reactivity for igg igm c3 and c1q of the glomeruli with a similar intensity. & the sample submitted for immunofluorescence comprises cortex and outer medulla. there is focal segmental mesangial and peripheral capillary wall positivity for igm 1+ c3 2+ and kappa <UNK> the significance of this is uncertain. reactions against igg iga c3 c1q kappa and lambda light chains and fibrinogen are all negative. there is no\\ 
\hline
\end{tabular}
\end{table}